\documentclass{article}
\usepackage{spconf,amsmath,amssymb,graphicx,booktabs,colortbl,xcolor}
\usepackage{placeins}
\usepackage{ragged2e}
\usepackage{stfloats}
\usepackage{hyperref}
\usepackage{capt-of}
\usepackage{caption}
\newcommand{\acr}[1]{{\Large #1}}
\title{\acr{CALM}: \acr{C}lass-wise \acr{A}greement and \acr{L}abel-Gated Disagreement \acr{M}odulation for Decentralized Federated Learning}

\name{Yifan Ying \qquad Qing Tian\sthanks{Corresponding author: qtian@uab.edu.}}
\address{Department of Computer Science, University of Alabama at Birmingham, Birmingham, AL, USA}

\begin{document}
\ninept
\maketitle

\begin{abstract}
Conventional federated learning relies on parameter averaging, which forces clients to be doubly homogeneous: all must run an identical architecture, and accuracy degrades when local data are non-IID. Decentralized federated distillation sidesteps both: each client runs its peers' model snapshots as teachers on its own local data and distills from their soft predictions, with no server, no public data, and no shared architecture. Under severe non-IID skew, however, the trustworthiness of the aggregated teacher target is a matter of degree, yet existing pipelines make hard, all-or-nothing decisions: outlier teachers are discarded by threshold, and whatever target survives is trusted in full. We propose CALM, which replaces every hard decision with a smooth trust gate at three levels: per class, teachers are weighted by agreement with the peer consensus; per sample, distillation is scaled by the teachers' divergence from that target; and a label gate scales it by how strongly the target supports the sample's true label. None of this adds communication or auxiliary data. On CIFAR-10, SVHN, OrganAMNIST, and Google Speech Commands with heterogeneous client architectures under Dirichlet label skew, CALM consistently outperforms uniform and hard-filtered distillation and matches or exceeds competing heterogeneous-FL methods.
\end{abstract}


\section{Introduction}
\label{sec:intro}

Federated learning (FL) trains models across clients without centralizing raw data \cite{McMahan2017ICML,Kairouz2021FL}, but its canonical instrument, parameter averaging, imposes two forms of homogeneity that practical deployments rarely satisfy: all clients must share one architecture, and performance degrades sharply when local data are non-IID \cite{Zhao2018NonIID,Yang2023Survey,Xu2024DFML,Jimenez2025NonIID}. Knowledge distillation (KD) escapes the architectural constraint by exchanging soft predictions rather than weights \cite{Hinton2015KD,Lin2020NeurIPS,Li2019FedMD}, and its fully decentralized variant removes the central server and public transfer set as well: each client shares only a model snapshot, instantiates its peers as local teachers, evaluates them on its own private data, and distills from their combined soft predictions \cite{Xu2024DFML,Banh2023DecentralizedKD}. The appeal is concrete: picture a fleet of voice-assistant devices jointly learning spoken-command recognition, where a home hub with an AI accelerator serves a full-scale acoustic model, a budget smartphone can host only compact ones, raw voice recordings must never leave the device, and each device hears its own skewed slice of the command vocabulary, speakers, and acoustic conditions. Peer distillation fits exactly: any two models over the same label space can exchange knowledge as black boxes, and because each teacher is evaluated on the student's own recordings, no audio ever leaves a device, yet knowledge of commands a device rarely hears still reaches it from peers that hear them often.

In this setting, the distillation target for every training sample is an aggregate of peer teacher predictions, and under heterogeneous, non-IID data its quality is sharply uneven: for a sample whose class most peers have barely seen, the teacher pool splinters into a few informed opinions and many guesses. Prior methods respond with hard decisions. Agreement filtering discards teachers that deviate from the peer consensus beyond a fixed threshold \cite{Bilbeisi2026CRAD}, echoing Byzantine-robust aggregation \cite{Blanchard2017NeurIPS}; the surviving teachers are then averaged, and the resulting target is imposed on the student with full force on every sample. Each hard decision loses graded information: a teacher just past the cutoff is silenced entirely while one just inside speaks at full volume, and a target formed from a pool in open conflict teaches as loudly as one backed by unanimous consensus. 

We propose to make trust graded everywhere. Our method, CALM (Class-wise Agreement and Label-gated Disagreement Modulation), replaces the pipeline's hard decisions with three smooth gates operating at three levels. At the teacher-class level, each teacher's contribution to the target for class $c$ is weighted by $\exp(-\Delta/\tau)$, where $\Delta$ is its deviation from the per-class peer consensus, so that agreement is rewarded continuously where a hard filter would impose a cutoff and discard the graded evidence. At the sample level, following the intuition that ensemble disagreement proxies epistemic uncertainty \cite{Lakshminarayanan2017NeurIPS}, the KD loss is scaled by $\exp(-d(x)/\sigma)$, where $d(x)$ is the teachers' mean divergence from the aggregated target, so contested targets contribute less than consensual ones. Both signals, however, are label-free and can be fooled when correlated teachers confidently agree on the wrong class. The third gate therefore anchors trust in the one piece of ground truth the client holds: distillation is further gated by $(\overline{q}^{\,y})^{\eta}$, the probability the aggregated target assigns to the sample's true label, so peer knowledge that contradicts local evidence is attenuated rather than absorbed.

Our contributions are: (1) we identify the all-or-nothing treatment of teacher trust, both in target construction and in loss application, as a failure mode of decentralized federated KD under non-IID data; (2) we propose CALM, three composable smooth trust gates (soft class-wise agreement, disagreement-modulated loss, and a label-consistency gate) that add no communication, no auxiliary data, and no learned parameters; and (3) on CIFAR-10, SVHN, OrganAMNIST, and Google Speech Commands, under heterogeneous architectures and severe Dirichlet skew, CALM improves global accuracy over uniform and hard-filtered distillation and matches or exceeds competing methods, with each gate contributing and the combination strongest. Beyond accuracy, the weight $\lambda(x,y)$ that CALM assigns to each sample also separates samples whose peer target is wrong from those whose target is right (Fig.~\ref{fig:lambda_separation}), a built-in indicator of where peer supervision can be trusted.

\begin{figure*}[t]
    \centering
    \includegraphics[width=0.925\textwidth]{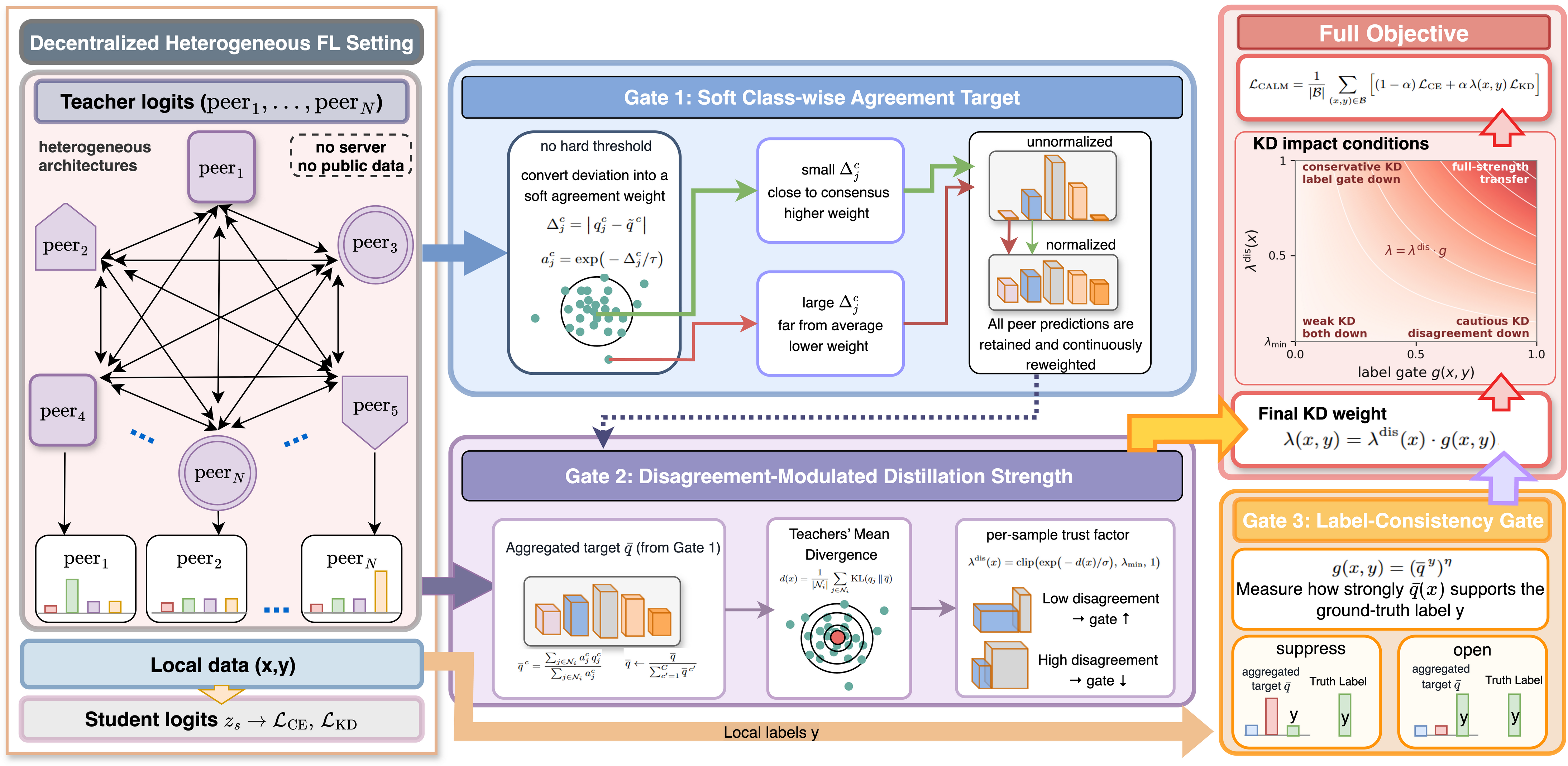}
    \caption{\protect\justifying
    Overview of the CALM framework.
Left: in a server-free, public-data-free setting, clients with heterogeneous architectures exchange model snapshots, never raw data, and a student obtains teacher logits by running each on its own local batch.
Top middle (Gate~1): CALM builds a soft class-wise agreement target, downweighting each teacher smoothly as it deviates from the peer consensus and renormalizing onto the probability simplex.
Bottom middle (Gate~2): teacher disagreement around the aggregated target becomes a per-sample trust factor that lowers distillation strength when the pool conflicts.
Bottom right (Gate~3): the label-consistency gate reads how strongly the target supports the ground-truth label, attenuating distillation when the target contradicts local evidence.
Top right: the final KD weight multiplies Gates~2 and~3; the heatmap shows $\lambda=\lambda^{\mathrm{dis}}\!\cdot g$ varying smoothly from full-strength to weak distillation, with no threshold anywhere.
}
\label{fig:overview}
\end{figure*}

\section{Relation to Prior Work}
\label{sec:prior}

\textbf{Heterogeneity in FL.} Parameter-space methods mitigate non-IID drift by regularization or control variates, e.g., FedProx \cite{Li2020MLSys} and SCAFFOLD \cite{Karimireddy2020ICML}, but operate in a shared weight space and cannot span distinct architectures \cite{Yang2023Survey}. KD-based FL exchanges predictions instead: FedMD \cite{Li2019FedMD} aligns clients on a public proxy set, generator-based methods synthesize the transfer set instead of collecting it \cite{zhang2023target}, and ensemble distillation \cite{Lin2020NeurIPS} fuses predictions on a central server that distills on its own public or generated data; all relocate the data dependency rather than remove it, or restore the central coordinator that FL set out to avoid. DFML \cite{Xu2024DFML} distills peer-to-peer with neither server nor public data, the regime we operate in. CRAD \cite{Bilbeisi2026CRAD} shares it, censoring teachers per class, whereas CALM replaces that hard decision with graded weighting and adds two sample-level gates.

\noindent\textbf{Weighting teachers and samples.} Existing methods weight teachers by data volume \cite{Lin2020NeurIPS}, by class-wise reliability \cite{wang2022knowledge}, or per sample by confidence against the ground-truth label \cite{Zhang2022CAMKD}, or censor outliers outright, echoing Byzantine-robust aggregation \cite{Blanchard2017NeurIPS}. These schemes decide which teachers shape the target and how strongly, yet even the label-aware among them assume co-located teachers and impose the fused target at full strength on every sample. CALM's agreement weighting is soft rather than censoring, and its two gates modulate how strongly each sample's fused target is imposed. Per-sample loss scaling has a long lineage---heteroscedastic attenuation \cite{Kendall2017NeurIPS}, curriculum learning \cite{Bengio2009ICML}---but unlike those unsupervised weightings, CALM also vets the target against the client's own labels.

\section{Method}
\label{sec:method}

Fig.~\ref{fig:overview} summarizes the framework. We consider $N$ clients with private datasets $\mathcal{D}_i$ over a shared label space $\{1,\dots,C\}$. Data are non-IID: class proportions differ sharply across clients, and a client may hold few or no samples of some classes. Models are heterogeneous: client $i$ runs an architecture $f_i$ suited to its hardware. There is no central server and no public dataset; the only interface any two clients share is the $C$-dimensional class posterior.

At communication round $r$, client $i$ receives its peers' round-$(r{-}1)$ model snapshots $\{\theta_j^{r-1}\}_{j\in\mathcal{N}_i}$, where the teacher set $\mathcal{N}_i\subseteq\{1,\dots,N\}\setminus\{i\}$ contains all peers or, under partial participation, the other clients active in the round. Client $i$ runs each snapshot locally as a teacher on its own training batches, so raw data never leaves the client. For a labeled sample $(x,y)\in\mathcal{D}_i$ the teacher soft predictions are
\begin{equation}
q_j = \mathrm{softmax}\!\left(f_{\theta_j^{r-1}}(x)/T\right)\in\Delta^{C-1},\quad j\in\mathcal{N}_i,
\label{eq:teacher_soft}
\end{equation}
with distillation temperature $T$ \cite{Hinton2015KD}. The conventional pipeline averages these uniformly, $\overline{q}=\frac{1}{|\mathcal{N}_i|}\sum_j q_j$ \cite{Lin2020NeurIPS}, optionally after censoring outlier teachers, and trains the student on
\begin{equation}
\mathcal{L}=(1-\alpha)\,\mathcal{L}_{\mathrm{CE}}+\alpha\,T^{2}\,\mathrm{KL}\!\left(\overline{q}\,\|\,\mathrm{softmax}(z_s/T)\right),
\label{eq:baseline_loss}
\end{equation}
where $z_s=f_i(x)$ and $\alpha\in[0,1]$ balances local supervision against distillation. Every sample's target is trusted equally in Eq.~\eqref{eq:baseline_loss}, and any filtering of teachers is binary. CALM removes both hard assumptions with three gates.

\subsection{Gate 1: Soft Class-wise Agreement Target}
\label{ssec:soft_target}

For each class $c$, we form the preliminary consensus (the uniform average of the conventional pipeline) and each teacher's deviation from it,
\begin{equation}
\tilde{q}^{\,c}=\frac{1}{|\mathcal{N}_i|}\sum_{j\in\mathcal{N}_i} q_j^{c},
\qquad
\Delta_j^{c}=\left|\,q_j^{c}-\tilde{q}^{\,c}\right|,
\label{eq:deviation}
\end{equation}
and convert deviation into a soft agreement weight,
\begin{equation}
a_j^{c}=\exp\!\big(-\Delta_j^{c}/\tau\big),
\label{eq:soft_agreement}
\end{equation}
with agreement scale $\tau$. The distillation target is the agreement-weighted average, renormalized onto the simplex:
\begin{equation}
\overline{q}^{\,c}=\frac{\sum_{j\in\mathcal{N}_i} a_j^{c}\, q_j^{c}}{\sum_{j\in\mathcal{N}_i} a_j^{c}},
\qquad
\overline{q}\leftarrow\frac{\overline{q}}{\sum_{c'=1}^{C}\overline{q}^{\,c'}}.
\label{eq:soft_target}
\end{equation}
This weighting subsumes hard agreement filtering without any hard cutoff: a teacher far from the consensus on class $c$ is exponentially attenuated rather than removed, and no teacher is ever irrevocably excluded. As $\tau\to 0$ the target concentrates on the most consensual teachers. Because the weights are computed per class, a teacher can be trusted for the classes it knows and discounted elsewhere, which is precisely the structure non-IID data induces.

\subsection{Gate 2: Disagreement-Modulated Distillation Strength}
\label{ssec:dmd}

Gate~1's target, however, carries only the teachers' relative deviations: the weighted average is unchanged when all agreement weights shrink by a common factor, precisely what happens when the entire pool drifts far from consensus. A pool in open conflict thus yields as clean-looking a target as a unanimous one, i.e., the severity of disagreement is erased in aggregation. We therefore reinstate that erased information as a per-sample signal, measuring how contested the target is by the teachers' mean divergence from it,
\begin{equation}
d(x)=\frac{1}{|\mathcal{N}_i|}\sum_{j\in\mathcal{N}_i}\mathrm{KL}\!\left(q_j\,\|\,\overline{q}\right),
\label{eq:disagreement}
\end{equation}
and convert it into a per-sample trust factor,
\begin{equation}
\lambda^{\mathrm{dis}}(x)=\mathrm{clip}\!\left(\exp\!\big(-d(x)/\sigma\big),\,\lambda_{\min},\,1\right),
\label{eq:modulation}
\end{equation}
where $\sigma$ sets the tolerance to disagreement and the floor $\lambda_{\min}$ prevents disagreement alone from silencing a sample, so that even contested samples retain a trickle of peer knowledge. Computing $d(x)$ reuses the $q_j$ already produced for aggregation, costing one divergence per teacher. Since the KD gradient is simply multiplied by $\lambda^{\mathrm{dis}}(x)$, this gate acts as a per-sample learning rate on borrowed knowledge: consensual targets teach at full strength, contested ones whisper.
\subsection{Gate 3: Label-Consistency Gate}
\label{ssec:label_gate}

Both gates above are label-free, and share a blind spot: teachers trained on similarly skewed shards can agree, confidently and unanimously, on the wrong class, in which case low disagreement certifies a bad target. The student, however, holds ground truth for its own training samples. We therefore read the aggregated target's mass on the true label $y$ as a direct check of its reliability and gate distillation by
\begin{equation}
g(x,y)=\left(\overline{q}^{\,y}\right)^{\eta},
\label{eq:label_gate}
\end{equation}
with exponent $\eta\ge 0$ controlling the gate's strength ($\eta{=}0$ disables it). If the peer target supports the true class, distillation proceeds; if it contradicts the local evidence, the sample's KD term is attenuated and learning falls back on cross-entropy supervision.

\subsection{Full Objective}
\label{ssec:objective}

The final per-sample distillation weight multiplies the two sample-level gates,
\begin{equation}
\lambda(x,y)=\lambda^{\mathrm{dis}}(x)\cdot g(x,y),
\label{eq:full_gate}
\end{equation}
and the training loss over a batch $\mathcal{B}\subset\mathcal{D}_i$ is
\begin{equation}
\mathcal{L}_{\mathrm{CALM}}
=\frac{1}{|\mathcal{B}|}\sum_{(x,y)\in\mathcal{B}}
\Big[(1-\alpha)\,\mathcal{L}_{\mathrm{CE}}
+\alpha\,\lambda(x,y)\,\mathcal{L}_{\mathrm{KD}}\Big],
\label{eq:calm_loss}
\end{equation}
where $\mathcal{L}_{\mathrm{KD}}=T^{2}\,\mathrm{KL}\!\left(\overline{q}\,\|\,\mathrm{softmax}(z_s/T)\right)$ and $\overline{q}$ is the soft agreement target of Eq.~\eqref{eq:soft_target}.
The three gates share one design principle: every trust decision in the pipeline, which teacher shapes the target for a class, how strongly a sample's target teaches, and whether it squares with the local label, is a smooth monotone function of the evidence rather than a threshold. Each gate degrades gracefully toward the baseline as its scale loosens ($\tau\to\infty$, $\sigma\to\infty$, $\eta\to0$), the gates compose by multiplication, and none adds communication, auxiliary data, or learned parameters.

\section{Experiments}
\label{sec:experiments}

\subsection{Setup}

\textbf{Datasets and partitioning.} We evaluate on four benchmarks spanning three modalities: CIFAR-10 and SVHN, natural-image benchmarks; OrganAMNIST, an eleven-class abdominal-CT organ-classification benchmark \cite{Yang2023MedMNIST} representing privacy-sensitive medical settings; and Google Speech Commands \cite{Warden2018Speech}, a spoken keyword-classification benchmark. We use a 10-class subset of Speech Commands v0.02; each waveform is resampled to 16 kHz and converted to a 32-bin mel spectrogram (\(n_{\mathrm{fft}}{=}512\), hop 160).
Each dataset is partitioned across all $N$ clients by a Dirichlet distribution with concentration $\alpha_{\mathrm{Dir}}{=}0.3$ (severe label skew); the skew-sensitivity study additionally uses $\alpha_{\mathrm{Dir}}\in\{0.1,0.5\}$.
The main comparison reports both $N{=}50$ and $N{=}100$, the latter doubling as the scaling study; the gate-by-gate ablation uses $N{=}50$.

\noindent\textbf{Compared methods.}\enspace We compare against FedMD \cite{Li2019FedMD}, FedGD \cite{zhang2023target}, MSFKD \cite{wang2022knowledge}, and DFML \cite{Xu2024DFML} (decentralized and public-data-free, the regime we share). FedMD and MSFKD use 5{,}000 unlabeled public proxy samples: CIFAR-100 images for CIFAR-10, SVHN-extra images for SVHN, PathMNIST images for OrganAMNIST, and 200 samples from each of 25 unused keywords for SpeechCommands; all are disjoint from client and test data. FedGD instead synthesizes 500 samples per round.
CALM uses neither. Within our backbone we compare: (i) Uniform KD, Eq.~\eqref{eq:baseline_loss} with uniform averaging; (ii) Hard filtering, per-class censoring of teachers with an adaptive median threshold; and (iii) full CALM.

\noindent\textbf{Federation and protocol.}\enspace Clients run three structurally distinct models assigned round-robin, ResNet-18 ($\approx$11M parameters), ResNet-18-Half ($\approx$2.8M), and CNN-6 ($\approx$0.8M), covering width and family heterogeneity. In each round, 10 uniformly drawn clients are active (modeling intermittent availability) and exchange snapshots peer-to-peer, so each teacher pool $\mathcal{N}_i$ is the other nine. We train for $R{=}300$ communication rounds with distillation weight $\alpha{=}0.7$, temperature $T{=}4$, agreement scale $\tau{=}0.1$, label-gate exponent $\eta{=}0.5$, disagreement scale $\sigma{=}1$, and floor $\lambda_{\min}{=}0.05$. All methods share the same client schedules, data partitions, and random seeds.
Checkpoints are selected on each client's local validation split and evaluated on the held-out global test set.

\subsection{Main Results}
\begin{table*}[!t]
\centering
\footnotesize
\caption{Comparison on CIFAR-10, SVHN, OrganAMNIST, and SpeechCommands with 50 total clients (left block) and 100 total clients (right block). Global accuracy (\%); best results are in bold.}
\label{tab:main}
\setlength{\tabcolsep}{12pt}
\begin{tabular}{lcccccccc}
\toprule
& \multicolumn{4}{c}{50 clients} & \multicolumn{4}{c}{100 clients} \\
\cmidrule(lr){2-5}\cmidrule(lr){6-9}
Method & CIFAR-10 & SVHN & OrganA& Speech & CIFAR-10 & SVHN & OrganA& Speech \\
\midrule
\multicolumn{9}{l}{\emph{Competing heterogeneous-FL methods}}\\
FedMD~\cite{Li2019FedMD}       & 20.26& 17.27& 20.25& 10.48& 10.50& 19.35& 20.34& 10.46\\
FedGD~\cite{zhang2023target}   & 21.01 & 10.12& 22.11& 17.15 & 11.38& 10.01& 13.47& 10.81 \\
MSFKD~\cite{wang2022knowledge} & 24.05& 16.08& 21.98& 19.09& 19.82& 18.31& 23.08& 10.85\\
DFML~\cite{Xu2024DFML}         & \textbf{40.07}& 70.35& 58.76& 61.49 & 29.43& 22.86& 46.98& 37.23\\
\midrule
\multicolumn{9}{l}{\emph{Our backbone, varying target trust}}\\
Uniform KD               & 28.43 & 44.24 & 53.51& 75.39 & 23.40& 23.78 & 50.24& 59.93 \\
Hard agreement filtering & 39.60& 58.57 & 59.38& 79.74 & 30.78& 34.87 & 51.13& 60.00 \\
\rowcolor{gray!20}
CALM (ours)              & 39.83 & \textbf{71.50} & \textbf{61.05}& \textbf{81.85} & \textbf{31.89} & \textbf{42.86} & \textbf{52.80}& \textbf{62.61} \\
\bottomrule
\end{tabular}
\end{table*}

\begin{table}[!t]
\centering
\footnotesize
\caption{Gate-by-gate ablation on SVHN with 50 total clients (10 active per round). Global accuracy (\%).}
\label{tab:components}
\setlength{\tabcolsep}{5pt}
\begin{tabular}{lcccc}
\toprule
Variant & Gate 1 & Gate 2 & Gate 3 & Global Acc. \\
\midrule
Uniform KD              & -- & -- & -- & 44.24\\
Soft target only        & \checkmark & -- & -- & 58.53\\
Soft target + disagreement & \checkmark & \checkmark & -- & 63.84\\
\rowcolor{gray!20}
CALM (ours)             & \checkmark & \checkmark & \checkmark & \textbf{71.50}\\
\bottomrule
\end{tabular}
\end{table}

\begin{table}[!t]
\centering
\footnotesize
\caption{Global accuracy (\%) on CIFAR-10 with 100 total clients (10 active per round) under increasing label skew. Smaller \(\alpha_{\mathrm{Dir}}\) means more skew.}
\label{tab:skew}
\setlength{\tabcolsep}{5pt}
\begin{tabular}{lccc}
\toprule
Method & \(\alpha_{\mathrm{Dir}}=0.1\) & \(\alpha_{\mathrm{Dir}}=0.3\) & \(\alpha_{\mathrm{Dir}}=0.5\) \\
\midrule
Uniform KD               & 20.36& 23.40& 27.59\\
Hard agreement filtering & 25.07& 30.78& 33.85\\
\rowcolor{gray!20}
CALM (ours)              & \textbf{26.21}& \textbf{31.89}& \textbf{35.46}\\
\bottomrule
\end{tabular}
\end{table}

Table~\ref{tab:main} reports global accuracy. CALM is best on every dataset under our backbone. It also consistently outperforms hard agreement filtering, with especially large gains on SVHN, and exceeds the strongest external baseline, DFML, on seven of the eight settings, trailing by only 0.24 points on 50-client CIFAR-10. Gains hold across all three modalities, so the gates are modality-agnostic. Scaling to 100 clients (right block) shrinks each shard and lowers every method that relies on local data, but CALM stays first on all four benchmarks (SVHN: 42.86\% vs.\ 23.78\% for uniform averaging).

\subsection{Analyses}

\noindent\textbf{Component ablation.}\enspace
Table~\ref{tab:components} builds CALM gate by gate on SVHN. The soft agreement target provides the largest single jump, lifting Uniform KD from 44.24\% to 58.53\%; disagreement modulation adds a further 5.3 points, and the label gate completes CALM at 71.50\%. The class-level agreement weighting (Gate~1) repairs the target by downweighting the few teachers that stray from the consensus, while the two complementary sample-level gates (Gates~2 and~3) discount the samples whose targets no reweighting can repair.
Disagreement modulation (Gate~2) damps contested, high-divergence targets, whereas the label gate (Gate~3) assigns systematically lower $g(x,y)$ to agreed-but-wrong targets that low disagreement alone leaves at full weight, which is why it yields the larger sample-level gain (+7.7 vs.\ +5.3 points).

\noindent\textbf{Robustness to skew.}\enspace
Table~\ref{tab:skew} varies the Dirichlet concentration $\alpha_{\mathrm{Dir}}\in\{0.1,0.3,0.5\}$ on CIFAR-10 (100 clients). CALM leads at every skew level, with the largest gain over uniform averaging at intermediate skew (+8.5 points at $\alpha_{\mathrm{Dir}}{=}0.3$); at extreme skew the margins narrow, as few teachers overlap each client's classes and little transferable signal remains to reweight.

\begin{figure}[!t]
    \centering

    \includegraphics[
        width=0.76\columnwidth
    ]{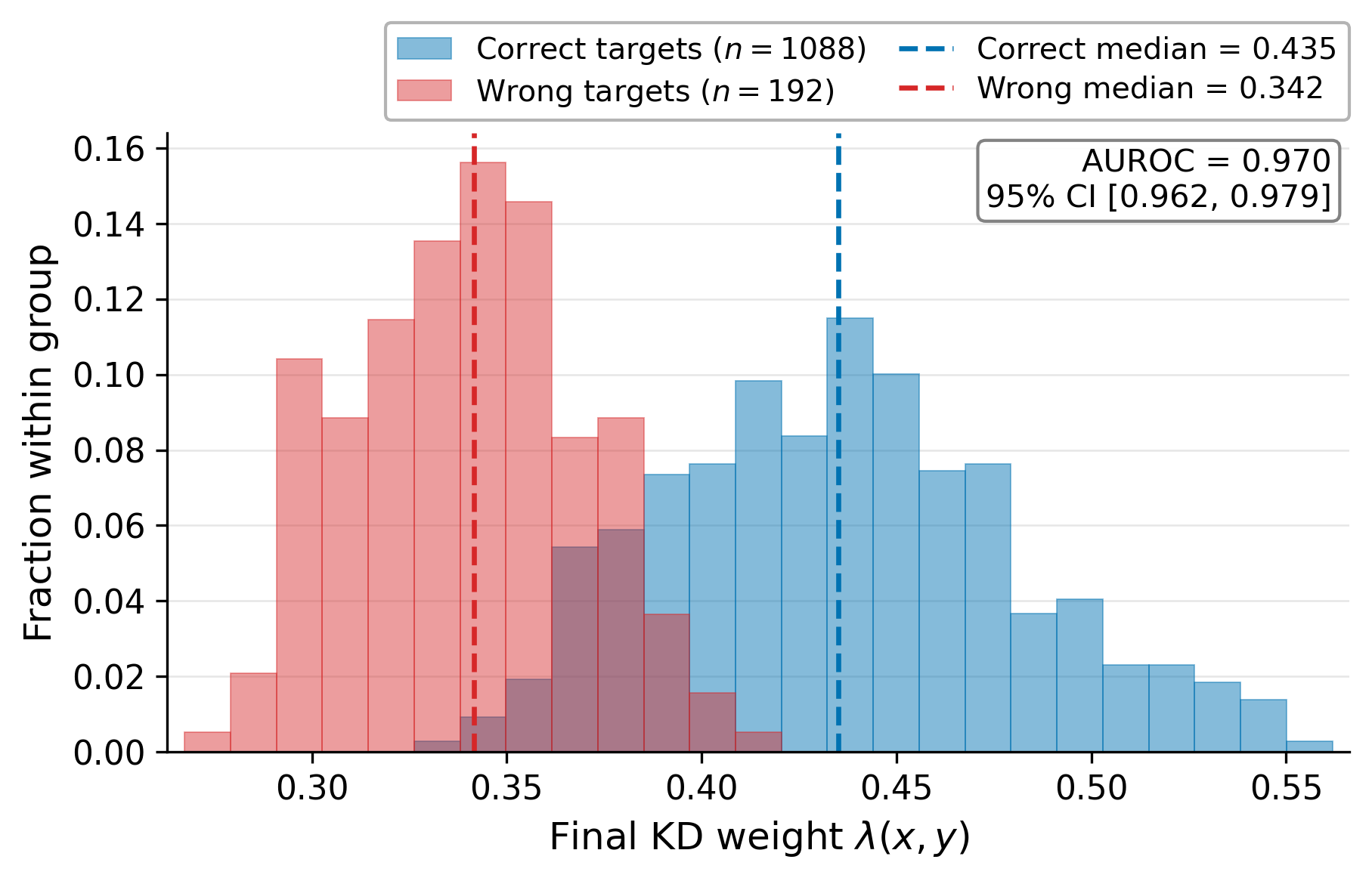}
    \caption{\protect\justifying
    Analysis of CALM's final KD weight $\lambda(x,y)$ on SVHN (50 clients, 10 active per round), final round.
    The distributions of $\lambda(x,y)$ for correct and incorrect
    aggregated targets separate clearly (AUROC 0.970, 95\%
    client-bootstrap CI $[0.962,\,0.979]$); histograms are normalized
    within group.
    }
    \label{fig:lambda_separation}
\end{figure}

\noindent\textbf{Trust-score reliability.}\enspace
Fig.~\ref{fig:lambda_separation} evaluates the deployed weight \(\lambda(x,y)\) as a per-sample trust score: wrong aggregated targets receive systematically lower \(\lambda\) (AUROC 0.970), so peer supervision is strongest where the target is right, and even correct targets are admitted only partially (median 0.435).

\section{Conclusion}
\label{sec:conclusion}

We introduced CALM for decentralized, public-data-free federated learning, replacing the pipeline's hard trust decisions with three smooth gates: class-wise agreement weighting builds the target, per-sample disagreement modulates how strongly it teaches, and a label gate anchors both in local labels. Under severe non-IID skew, CALM outperforms uniform distillation and hard filtering, and matches or exceeds prior decentralized KD.


\begin{thebibliography}{10}

\bibitem{McMahan2017ICML}
H.~Brendan McMahan, Eider Moore, Daniel Ramage, Seth Hampson, and Blaise~Ag{\"u}era y~Arcas,
\newblock ``Communication-efficient learning of deep networks from decentralized data,''
\newblock in {\em Proceedings of the 20th International Conference on Artificial Intelligence and Statistics (AISTATS)}, 2017, vol.~54 of {\em Proceedings of Machine Learning Research}, pp. 1273--1282.

\bibitem{Kairouz2021FL}
Peter Kairouz, H.~Brendan McMahan, Brendan Avent, Aur{\'e}lien Bellet, Mehdi Bennis, Arjun~Nitin Bhagoji, Keith Bonawitz, Zachary Charles, Graham Cormode, Rachel Cummings, Rafael G.~L. D'Oliveira, Salim~El Rouayheb, David Evans, Josh Gardner, Zachary Garrett, Adri{\`a} Gasc{\'o}n, Badih Ghazi, Phillip~B. Gibbons, Marco Gruteser, Za{\"i}d Harchaoui, et~al.,
\newblock ``Advances and open problems in federated learning,''
\newblock {\em Foundations and Trends in Machine Learning}, vol. 14, no. 1--2, pp. 1--210, 2021.

\bibitem{Zhao2018NonIID}
Yue Zhao, Meng Li, Liangzhen Lai, Naveen Suda, Damon Civin, and Vikas Chandra,
\newblock ``Federated learning with {Non-IID} data,''
\newblock {\em arXiv preprint arXiv:1806.00582}, 2018.

\bibitem{Yang2023Survey}
Boyu Fan, Siyang Jiang, Xiang Su, Sasu Tarkoma, and Pan Hui,
\newblock ``A survey on model-heterogeneous federated learning: Problems, methods, and prospects,''
\newblock in {\em Proceedings of the IEEE International Conference on Big Data (BigData)}, 2024, pp. 7725--7734.

\bibitem{Xu2024DFML}
Yasser~H. Khalil, Amir~Hossein Estiri, Mahdi Beitollahi, Nader Asadi, Sobhan Hemati, Xu~Li, Guojun Zhang, and Xi~Chen,
\newblock ``{DFML}: Decentralized federated mutual learning,''
\newblock {\em Transactions on Machine Learning Research}, 2024.

\bibitem{Jimenez2025NonIID}
Daniel~M. Jimenez-Gutierrez, Mehrdad Hassanzadeh, Aris Anagnostopoulos, Ioannis Chatzigiannakis, and Andrea Vitaletti,
\newblock ``A thorough assessment of the non-{IID} data impact in federated learning,''
\newblock {\em Journal of Industrial Information Integration}, vol. 50, 2026,
\newblock Art. no. 101052.

\bibitem{Hinton2015KD}
Geoffrey Hinton, Oriol Vinyals, and Jeff Dean,
\newblock ``Distilling the knowledge in a neural network,''
\newblock {\em arXiv preprint arXiv:1503.02531}, 2015.

\bibitem{Lin2020NeurIPS}
Tao Lin, Lingjing Kong, Sebastian~U. Stich, and Martin Jaggi,
\newblock ``Ensemble distillation for robust model fusion in federated learning,''
\newblock in {\em Advances in Neural Information Processing Systems 33 (NeurIPS)}, 2020, pp. 2351--2363.

\bibitem{Li2019FedMD}
Daliang Li and Junpu Wang,
\newblock ``{FedMD}: Heterogeneous federated learning via model distillation,''
\newblock {\em arXiv preprint arXiv:1910.03581}, 2019.

\bibitem{Banh2023DecentralizedKD}
Eunjeong Jeong and Marios Kountouris,
\newblock ``Personalized decentralized federated learning with knowledge distillation,''
\newblock in {\em IEEE International Conference on Communications (ICC)}, 2023, pp. 1982--1987.

\bibitem{Bilbeisi2026CRAD}
Baraa Bilbeisi, Mengchen Fan, Baocheng Geng, and Qing Tian,
\newblock ``{CRAD}: Class-wise reliability-aware distillation for decentralized heterogeneous federated learning,''
\newblock {\em arXiv preprint arXiv:2609.00446}, 2026.

\bibitem{Blanchard2017NeurIPS}
Peva Blanchard, El~Mahdi~El Mhamdi, Rachid Guerraoui, and Julien Stainer,
\newblock ``Machine learning with adversaries: Byzantine tolerant gradient descent,''
\newblock in {\em Advances in Neural Information Processing Systems 30 (NIPS)}, 2017, pp. 119--129.

\bibitem{Lakshminarayanan2017NeurIPS}
Balaji Lakshminarayanan, Alexander Pritzel, and Charles Blundell,
\newblock ``Simple and scalable predictive uncertainty estimation using deep ensembles,''
\newblock in {\em Advances in Neural Information Processing Systems 30 (NIPS)}, 2017, pp. 6402--6413.

\bibitem{Li2020MLSys}
Tian Li, Anit~Kumar Sahu, Manzil Zaheer, Maziar Sanjabi, Ameet Talwalkar, and Virginia Smith,
\newblock ``Federated optimization in heterogeneous networks,''
\newblock in {\em Proceedings of Machine Learning and Systems (MLSys)}, 2020, vol.~2, pp. 429--450.

\bibitem{Karimireddy2020ICML}
Sai~Praneeth Karimireddy, Satyen Kale, Mehryar Mohri, Sashank~J. Reddi, Sebastian~U. Stich, and Ananda~Theertha Suresh,
\newblock ``{SCAFFOLD}: Stochastic controlled averaging for federated learning,''
\newblock in {\em Proceedings of the International Conference on Machine Learning (ICML)}, 2020, pp. 5132--5143.

\bibitem{zhang2023target}
Jie Zhang, Chen Chen, Weiming Zhuang, and Lingjuan Lyu,
\newblock ``{TARGET}: Federated class-continual learning via exemplar-free distillation,''
\newblock in {\em Proceedings of the IEEE/CVF International Conference on Computer Vision (ICCV)}, 2023, pp. 4759--4770.

\bibitem{wang2022knowledge}
Dong Wang, Naifu Zhang, Meixia Tao, and Xu~Chen,
\newblock ``Knowledge selection and local updating optimization for federated knowledge distillation with heterogeneous models,''
\newblock {\em IEEE Journal of Selected Topics in Signal Processing}, vol. 17, no. 1, pp. 82--97, 2023.

\bibitem{Zhang2022CAMKD}
Hailin Zhang, Defang Chen, and Can Wang,
\newblock ``Confidence-aware multi-teacher knowledge distillation,''
\newblock in {\em Proceedings of the IEEE International Conference on Acoustics, Speech and Signal Processing (ICASSP)}, 2022, pp. 4498--4502.

\bibitem{Kendall2017NeurIPS}
Alex Kendall and Yarin Gal,
\newblock ``What uncertainties do we need in bayesian deep learning for computer vision?,''
\newblock in {\em Advances in Neural Information Processing Systems 30 (NIPS)}, 2017, pp. 5574--5584.

\bibitem{Bengio2009ICML}
Yoshua Bengio, J{\'e}r{\^o}me Louradour, Ronan Collobert, and Jason Weston,
\newblock ``Curriculum learning,''
\newblock in {\em Proceedings of the International Conference on Machine Learning (ICML)}, 2009, pp. 41--48.

\bibitem{Yang2023MedMNIST}
Jiancheng Yang, Rui Shi, Donglai Wei, Zequan Liu, Lin Zhao, Bilian Ke, Hanspeter Pfister, and Bingbing Ni,
\newblock ``{MedMNIST} v2 -- a large-scale lightweight benchmark for {2D} and {3D} biomedical image classification,''
\newblock {\em Scientific Data}, vol. 10, no. 1, 2023,
\newblock Art. no. 41.

\bibitem{Warden2018Speech}
Pete Warden,
\newblock ``Speech commands: A dataset for limited-vocabulary speech recognition,''
\newblock {\em arXiv preprint arXiv:1804.03209}, 2018.

\end{thebibliography}
\end{document}